\documentclass[preprint,12pt]{elsarticle}

\usepackage{amssymb}

\usepackage{amsmath, graphicx, booktabs, tabularx, multirow, pgfplots, tabu, caption, subcaption, bm, float}
\usepackage{marvosym}
\usepackage{ifsym}
\usepackage{threeparttable}
\usepackage{etoolbox}
\patchcmd{\thebibliography}
{\settowidth}
{\setlength{\parsep}{0pt}\setlength{\itemsep}{0pt plus 0.1pt}\settowidth}
{}{}
\usepackage[colorlinks, linkcolor=red, citecolor=blue]{hyperref}
\usepackage{array}

\journal{Information Sciences}

\begin{document}
	
	\begin{frontmatter}
		
		
		
		\title{Ada-DF: An Adaptive Label Distribution Fusion Network For Facial Expression Recognition}
		
		
		\author[label1]{Shu Liu, Yan Xu, Tongming Wan, Xiaoyan Kui}
		
		\affiliation[label1]{organization={School of Computer Science and Engineering},
			addressline={Central South University}, 
			city={Changsha},
			postcode={410083}, 
			state={Hunan},
			country={China}}
		
		\renewcommand{\thefootnote}{\fnsymbol{footnote}}
		
		\begin{abstract}
			
			Facial expression recognition (FER) plays a significant role in our daily life. However, annotation ambiguity in the datasets could greatly hinder the performance. In this paper, we address FER task via label distribution learning paradigm, and develop a dual-branch Adaptive Distribution Fusion (Ada-DF) framework. One auxiliary branch is constructed to obtain the label distributions of samples. The class distributions of emotions are then computed through the label distributions of each emotion. Finally, those two distributions are adaptively fused according to the attention weights to train the target branch. Extensive experiments are conducted on three real-world datasets, RAF-DB, AffectNet and SFEW, where our Ada-DF shows advantages over the state-of-the-art works. \footnote{The code is available at \url{https://github.com/taylor-xy0827/Ada-DF}.}
			
		\end{abstract}
		
		%
		
		\begin{keyword}
			Facial Expression Recognition \sep Adaptive Distribution Fusion \sep Label Distribution Learning \sep Dual Branch
			
			
		\end{keyword}
		
	\end{frontmatter}
	
	
	\section{Introduction}
	\label{sec:intro}
	Facial expression plays a pivotal role in human communication, which serves as a crucial medium for conveying emotions. In the realm of affective computing, automatic facial expression recognition (FER) has found extensive applications in many fields such as psychotherapy \citep{candra2016classification}, remote teaching \citep{savchenko2022classifying}, etc.
	
	Notably, recent advancements in deep learning coupled with the availability of large-scale datasets have made great progress in FER \citep{li2020deep}, surpassing the performance of traditional methods. Deep learning methods, including convolutional neural networks (CNNs) \citep{gu2018recent} and recurrent neural networks (RNNs) \citep{medsker2001recurrent}, have revolutionized the filed by effectively capturing intricate facial features, consequently enhancing the accuracy and robustness of FER systems. Moreover, many comprehensive FER datasets encompassing diverse facial expressions captured under various conditions have emerged, which facilitate the training of deep learning models, enable the improved generalization and refine the real-world applicability. The integration of deep learning and large-scale datasets has opened up new avenues for FER methods, holding immense potential for understanding and interpreting human emotions in a wide range of practical contexts. 
	
\begin{figure}
		\centering
		\includegraphics[width=0.5\textwidth]{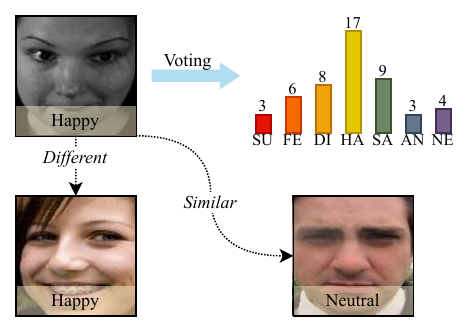}
		\caption{Annotations by 50 volunteers on an image from RAF-DB. The face is annotated with 7 basic emotions, including \textbf{su}rprise, \textbf{fe}ar, \textbf{di}sgust, \textbf{ha}ppiness, \textbf{sa}dness, \textbf{an}ger and \textbf{ne}utral. Another two faces are chosen to show the inter-class similarity and intra-class variation. }
		\label{fig:example}
	\end{figure}

	However, the performance of deep-learning-based FER methods is often hindered by the inherent ambiguity of real-world FER datasets. Comparing to facial images captured in the lab, human faces in the wild exhibit variations in illumination, head pose and identity, making it challenging to capture consistent and reliable facial expressions. These expression-unrelated variations introduce ambiguity that is difficult to mitigate. More attention are paid to the ambiguity introduced by uncertain annotations. Most datasets are annotated with 7 basic emotions \citep{ekman1999basic} by multiple volunteers, and follow the majority voting scheme to assign the most chosen single label to each sample. Nevertheless, these annotations often fall short of meeting the actual requirements for FER due to the unprofessional labeling caused by the subjectiveness of annotators. Furthermore, recent studies have demonstrated that individuals from different cultural and social backgrounds tend to categorize the same facial image into different emotions \citep{ekman1971universals}, resulting in both intra-class similarity and inter-class variation. To illustrate this point, we randomly selected an image from the RAF-DB \citep{li2017reliable} dataset and invited 50 volunteers with diverse genders, ages, and nationalities to annotate the image shown in Figure \ref{fig:example}. The original annotation for this image was \textit{happy}. Although a majority of volunteers identified it as \textit{happy}, the total number of volunteers choosing \textit{sad} and \textit{disgusted} was comparable to those selecting \textit{happy}, indicating that assigning a single label to this sample is not appropriate. Moreover, this face image looks quite different from another face annotated with \textit{happy}, but similar to the face annotated with \textit{neutral}, showing the great inter-class variation and intra-class similarity. 
	
	To address the ambiguity problem in FER, the label distribution learning (LDL) \citep{geng2016label} is introduced, which assigning different weights to all emotions. However, LDL-based methods still face challenges in obtaining high-quality label distribution annotations, further exacerbating the ambiguity problem. In this paper, we present a straightforward yet effective approach to address the challenges in FER by focusing on two key aspects: class distribution mining and label distribution fusion. For the former, an auxiliary branch is constructed to extract the label distributions of samples. Due to the ambiguity in those samples, these distributions lack the accuracy to describe facial features in the label space. To overcome this limitation, the class distribution mining is introduced to exclude the biases existing in the label distributions and mine the rich sentiment behind each emotion. For the latter, we design an adaptive distribution fusion module that leverages attention weights to effectively merge the label distributions of samples and the class distributions of emotions. 
	
	The main contributions of our research can be summarized as follows:
	
	(1) We present a multi-task LDL framework which can discriminate ambiguous or mislabeled samples jointly by attention module while optimizing label distribution generation on the auxiliary branch and facial expression classification on the target branch. 
	
	(2) We propose a class distribution mining module which extracts the class distributions of emotions for each class of expressions from the label distributions of samples output by the auxiliary branch, thus excluding biases in the label distributions and mining the rich sentiment information behind each class of expressions. 
	
	(3) We propose an adaptive distribution fusion module which adaptively fuse the label distributions of samples and the class distributions of emotions according to ambiguity degrees of samples to obtain the fused distributions with more accuracy, thus providing more accurate and richer supervision for training the model.  
	
	\section{Related Work}
	\label{sec:related}
	
	\subsection{Facial Expression Recognition}
	In recent years, numerous FER methods have emerged, mostly encompassing several essential steps: data acquisition, feature extraction, and expression classification, which collectively contribute to the overall FER pipeline. 
	
	FER methods mostly rely on public FER datasets for training, such as JAFFE \citep{lyons2021excavating}, CK+ \citep{lucey2010extended} from controlled lab environments. While these in-the-lab datasets promote the early FER research, their limitations arise in real-world scenarios where face images exhibit variations in illumination, pose and occlusion. Consequently, models trained on these datasets struggle to perform and generalize well in practical applications. However, with the rapid development of internet, large real-world datasets based on online face images have emerged, such as RAF-DB and AffectNet \citep{mollahosseini2017affectnet}, which offer a more diverse and representative collection of facial expressions, greatly boosting the performance of deep learning-based FER methods. Additionally, face detectors, such as MTCNN (Multi-Task Cascaded Convolutional Networks) \citep{xiang2017joint} and RetinaFace \citep{deng2020retinaface}, are commonly deployed for facial alignment to localize faces in real-world face images. 
	
	Based on different extracted features, FER methods can be roughly divided into handcrafted features and deep learning-based methods. Handcrafted features mainly focus on texture information of face images, such as local binary patterns (LBP) \citep{ahonen2004face} and histogram of oriented gradients (HOG) \citep{dalal2005histograms}. On the other hand, motivated by CNNs and large-scale datasets, deep learning-based methods can automatically learn expressive features from face images. While handcrafted features-based methods have computational efficiency and strong interpretability, deep learning-based methods have shown superior performance and generalization with more computation cost and time consumption. Thanks to the rapid advancement of computing hardware, especially GPUs, deep learning-methods can be more easily trained and employed. 
	
	Expression classification serves as the final step in a FER system. Once the features are extracted, they are input into a classifier which assigns the corresponding emotion label to each face image. Classification algorithms, including support vector machines (SVMs) \citep{gunn1998support}, decision trees \citep{safavian1991survey}, or CNNs, are commonly employed. Specially, deep learning-based methods can perform in an end-to-end manner, which means the model classifies samples and optimizes itself without explicit feature extraction, eliminating the complex manual feature engineering and potentially improving the overall FER performance. 
	
	\subsection{Label distribution learning}
	Previous learning paradigms, such as single label-learning and multi label-learning, can hardly satisfy real applications. Motivated by existing data with various importance, a new learning paradigm called label distribution learning (LDL) was proposed in 2016, which replaces single labels with different weights of labels. Comparing to single labels, label distributions contain richer information and can effectively mitigate overfitting, thus providing better supervision for model training. However, most methods still rely on single label-learning because it is challenging to obtain label distributions as most datasets are only annotated with single labels. 
	
	To overcome the difficulty of obtaining label distributions, two simple but effective approaches are employed by recent LDL methods: constructing Gaussian distributions and using probability distributions. The former assumes that the actual label distributions of samples follow the Gaussian distributions centered around the original labels of samples, commonly used in regression tasks such as age prediction \citep{rothe2018deep}, and occasionally in few classification tasks like image emotion recognition \citep{yang2017joint}. The latter directly utilizes the probability distributions output by the model as label distributions, commonly used in classification tasks such as object recognition \citep{gao2021learning}. As facial expression recognition can be viewed as an image classification task, we adopt the latter approach to extract the label distributions.
	
	\subsection{Learning against Uncertainties}
	Though deep learning has significantly improved FER performance, deep learning-methods still suffer from the inherent ambiguity of FER datasets, which can be divided into data ambiguity and label ambiguity. Data ambiguity is introduced by various factors, including illuminations, head poses and identity biases, which greatly affect the data quality. To address these issues, facial alignment and data cleaning \citep{rahm2000data} are commonly employed to reduce the impact of FER-unrelated factors. On the other hand, label ambiguity primarily arises from discrete annotations which are against the consumption that individuals under different backgrounds tend to categorize facial expressions differently. Furthermore, FER datasets often exist inter-class similarity and intra-class variation, making it hard to classify an expression accurately.
	
	Many recent methods have been proposed to cope with label ambiguity. Semi-supervised learning (SSL) methods have shown promising results, especially pseudo-labeling, a prominent SSL approach assigning labels to unlabeled data based on model predictions. Sohn etc. \citep{sohn2020fixmatch} proposed a SSL framework, which acquired pseudo labels base on a fixed threshold. Furthermore, Li etc. \citep{li2022towards} further boosted SSL performance via adaptive confidence learning which set unique threshold for each emotion.
	
	In recent years, LDL-based methods have been raised to cope with label ambiguity as label distributions have richer information than single labels, thus mining the relationships between labels and eliminating the ambiguity. DMUE \citep{she2021dive} generated the label distributions of samples by constructing multiple auxiliary branches which corresponded to different classes of expressions respectively, excluding interference from samples of different classes. LDL-ALSG \citep{chen2020label} improved the performance by minimizing the distance between the logical label and the label distribution of a sample, and the distance between the label distribution of a samples and the label distributions of its neighbors in the label space of action unit recognition and facial landmark detection. Motivated by these previous works, our methods adopts LDL as the overall learning paradigm. 
	
	\begin{figure*}
		\centering
		\includegraphics[width=\textwidth]{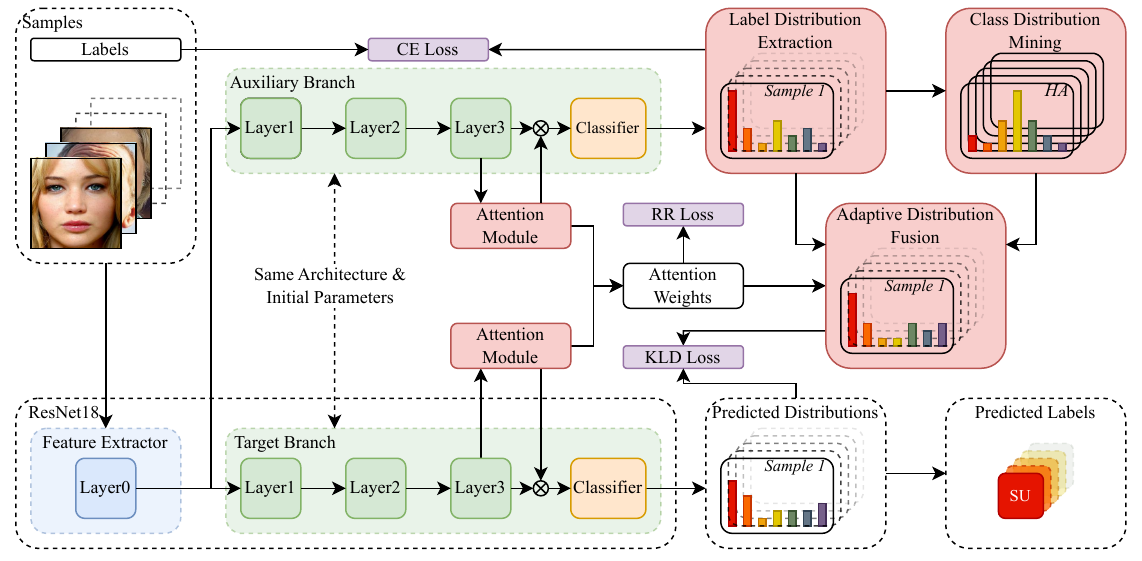}
		\caption{\textbf{Ada-DF} contains an auxiliary branch for label distribution generation, and a target branch for final emotion prediction.  The attention weights of two attention modules are normalized to integrate the label distributions and the class distributions. The final fused distributions are used to train the target branch via label distribution learning.}
		\label{fig:model}
	\end{figure*}
	
	\section{Method}
	\label{sec:method}
	
	We propose a novel dual-branch adaptive distribution fusion network Ada-DF to tackle the ambiguity problem in FER, as shown in Figure \ref{fig:model}. In this section, we first introduce how we extract the label distributions of samples. Then, we explain the class distribution mining and dynamic distribution fusion in detail. Finally, the joint multi-task loss is introduced, which optimizes the overall framework. 
	
	\subsection{Label Distribution Extraction}
	
	According to convention, we typically use the probability distributions output by the model as the label distributions of samples. However, directly training the model using the probability distributions can lead to performance degradation. To address this, we construct an independent auxiliary branch to extract the label distributions of samples. The details of the auxiliary branch are as follows: First, we pretrain our model on the MS-Celeb-1M \citep{guo2016ms} dataset following the convention, enabling the model to benefit from transferred knowledge in face recognition. Then, we divide the model into two parts: freezing the first few convolutional layers as the feature extractor, and using the latter few convolutional layers as the feature discriminator. Finally, we construct an auxiliary branch with parameters and structure consistent with the feature discriminator to extract the label distributions of samples. We refer to the feature discriminator in the original model as the target branch, corresponding to the auxiliary branch. The process of constructing an auxiliary branch is applicable to all deep learning-based models, such as deep neural networks, visual transformers, and so on. Specially, our auxiliary branch construction is based on the ResNet18 \citep{he2016deep} backbone network, with the feature extractor composed of the first convolutional layer and the feature discriminator composed of the last three convolutional layers of ResNet18. The label distributions of samples can be formulated as follows:
	
	\begin{equation}
		\mathbf{d}_{label, x_i} = \{d_{label, x_i}^{y_j}, j = 1, 2, \dots, C\}
	\end{equation}	
	with
	\begin{equation}
		d_{label, x_i}^{y_j}=p_{aux, x_i}^{y_j}\left(x_i, \theta_{aux}\right)
	\end{equation}
	where $y_j$ is the $j$-th label, $d_{label, x_i}^{y_j}$ is the description degree of $y_j$ to sample $x_i$, $p_{aux, x_i}^{y_j}\left(x_i, \theta_{aux}\right)$ is the prediction probability of sample $x_i$ for class $y_j$ by the auxiliary branch. 
	
	The auxiliary branch will keep training via the cross-entropy loss to enhance and maintain the capability of the auxiliary branch in extracting the label distributions of samples. The loss function is formulated as follows:
	
	\begin{equation}
		L_{ce} = -\frac{1}{N}\sum_{i=1}^N\sum_{c=1}^Cy_i^c \log{\left(p_{aux, x_i}^c\left(x_i, \theta_{aux}\right)\right)}
	\end{equation}
	
	where $y_j$ is the $j$-th label, $p_{aux, x_i}^{y_j}\left(x_i, \theta_{aux}\right)$ is the prediction probability of sample $x_i$ for class $y_j$ by the auxiliary branch. 
	
	\subsection{Class Distribution Mining}
	
	Due to the sensitivity of deep neural networks to ambiguous or mislabeled samples, we employ class distribution mining to identify the underlying invariants within the label distributions of samples. This module aims to mitigate the impact of distribution errors on model performance. The class distributions of emotions are formulated as follows:
	
	\begin{equation}
		\mathbf{d}_{class}^c=\frac{\sum_{i=1}^{N_c}d_{label, x_i}^c}{N_c}, c = 1, 2, \dots, C
	\end{equation}
	
	where $\mathbf{d}_{class}^c$ is the class distribution of class $c$, $N_c$ is the number of samples belonging to class $c$.  
	
	During the initial epochs of training, the parameters of the auxiliary branch may be unstable, leading to an inability to output robust class distributions that accurately describe each facial expression category. To prevent erroneous class distributions from degrading the predictive performance of the model, we introduce a threshold $t$ to assess the stability of the class distributions at each epoch. If the description degree for a particular category does not meet the threshold, we temporarily replace the class distribution with a threshold distribution for model training. This approach ensures that the model receives more reliable and consistent supervision, preventing the negative impact of unstable class distributions on the training process. The improved class distributions of emotions are formulated as follows:
	
	\begin{equation}
		\mathbf{d}_{class}^c = 
		\begin{cases}
			\frac{\sum_{i=1}^{N_c}d_{label, x_i}^c}{N_c} & d_{class}^{c, y_c} \ge t \\
			\mathbf{d}_{thre}^c & d_{class}^{c, y_c} < t
		\end{cases}
	\end{equation}
	with
	\begin{equation}
		\mathbf{d}_{thre}^c = \{d_{thre}^{c, y_j}, j=1, 2, \dots, C\}
	\end{equation}
	with
	\begin{equation}
		d_{thre}^{c, y_j} = 
		\begin{cases}
			t & j=c \\
			\frac{1-t}{C-1} & j \neq c
		\end{cases}
	\end{equation}
	where $d_{thre}^{c, y_j}$ is the description degree of $y_j$ for the threshold distribution corresponding to class $c$.
	
	\subsection{Dynamic Distribution Fusion}
	
	In our preliminary experiments, both label distributions $\mathbf{d}_{label}$ and class distributions $\mathbf{d}_{class}$ achieve good results on benchmark datasets. However, they still cannot exceed SOTA due to unstable label distribution generation and intra-class variations. To further boost the model's performance, we try to solve the ambiguity problem via dynamic distribution fusion. 
	
	Dynamic distribution fusion is based on class distributions of emotions and adaptively combines them with label distributions of samples according to attention weights of each sample. Therefore, adaptive distribution fusion can be divided into two steps: attention weights extraction and adaptive distribution fusion. 
	
	\subsubsection{Attention Weights Extraction}
	
	To obtain the attention weights of samples, we embed two attention modules into the last layer of each branch. The attention module consists of a fully connected layer and the sigmoid function. The overall process of attention weights extraction is as follows: First, for a batch of samples, we input the facial features extracted by the feature extractor into both the auxiliary branch and the target branch. Next, the features output by each branch are input into their respective attention modules to extract the attention weights for each samples. The features from each branch are then multiplied by their corresponding attention weights and input into their respective classifiers. We calculate the average attention weights from both branches to benefit from the ambiguity discernment of both branches. The averaged attention weights can be represented as follows:
	
	\begin{equation}
		w_{avg, x_i} = \frac{w_{aux, x_i}+w_{tar, x_i}}{2}
	\end{equation}
	
	where $w_{aux, x_i}$ and $w_{tar, x_i}$ are attention weights of the $i$-th sample from two attention modules. The rank regularization is then used to adjust the attention weights to avoid the degradation of discernment ability in both branches:
	
	\begin{equation}
		L_{RR} = \max{(0, \delta - \left(w_H - w_L\right))}
	\end{equation}
	with
	\begin{equation}
		w_H = \frac{1}{M}\sum_{i=1}^Mw_{avg, x_i}, w_L = \frac{1}{N-M}\sum_{i=M+1}^Nw_{avg, x_i}
	\end{equation}
	
	where $\delta$ is a fixed margin, $w_H$ and $w_L$ are the mean values of the high weight group with $M$ samples and the low weight group with $N-M$ samples. For simplicity, values of $\delta$ and $M$ are directly adopted from SCN \citep{wang2020suppressing}, which are $0.07$ and $0.7N$ respectively. Finally, we normalize the attention weights to $w_{min}$ ~ 1 to ensure that the sum of description degrees for each class distribution is equal to 1 during adaptive distribution fusion:
	
	\begin{equation}
		\mathbf{w} = \frac{\mathbf{w}_{avg}-\min{(\mathbf{w}_{avg})}}{\max{(\mathbf{w}_{avg})}-\min{(\mathbf{w}_{avg})}} \cdot \left(1-w_{min}\right)+w_{min}
	\end{equation}
	
	where $w_{min}$ is the value of the lower limit for the attention weights after normalization. 
	
	\subsubsection{Adaptive Distribution Fusion}
	
	After obtaining the attention weights of samples, we employ adaptive distribution fusion to balance the robustness of class distributions of emotions and the diversity of label distributions of samples. The fused distributions are formulated as follows:
	
	\begin{equation}
		\mathbf{d}_{fused, x_i} = w_{x_i} \cdot \mathbf{d}_{class, x_i}^{y_i} + \left(1-w_{x_i}\right) \cdot \mathbf{d}_{label, x_i}
	\end{equation}
	
	where $d_{fused, x_i}$ is the fused distribution of sample $x_i$. 
	
	We then explain why adaptive distribution is effective. Due to the presence of ambiguous or mislabeled samples, the label distributions of samples output by the model often fluctuates around their ground-truth distributions. We assume that the fluctuations follow the normal distribution and use class distribution mining to eliminate those biases introduced by the fluctuations. However, class distributions are only able to accurately describe the emotions contained in certain samples since they are directly associated with the original single labels, which means that class distributions of emotions cannot properly describe ambiguous samples. Therefore, we need to balance the label distributions of samples and the class distributions of emotions to obtain the actual distributions. 
	
	We define the difference between the class distribution and the label distribution as the relative distribution of a sample. For samples with high attention weights, which are often clear samples, their relative distributions mainly represents the biases introduced by the fluctuations. Therefore, we assign relatively lower weights to the relative distributions of these samples. By directly using the class distributions to train these samples, we can achieve good results. For samples with low attention weights, which are often ambiguous or mislabeled samples, their relative distributions differs significantly from the that of clear samples. In this case, the differences of emotion information contained in these relative distributions are much larger than the biases. Thus, we assign relatively higher weights to the relative distributions of samples. The fused distributions are closer to the ground-truth distributions of the samples, helping the model better learn from ambiguous samples. 
	
	The target branch is trained by the Kullback-Leibler divergence loss  between the fused distributions and the predicted distributions as follows:
	\begin{equation}
		L_{kld} = -\frac{1}{N}\sum_{i=1}^N\sum_{c=1}^Cd_{fused, x_i}^c\log{\left(\frac{d_{fused, x_i}^c}{p_{tar, x_i}^c\left(x_i, \theta_{tar}\right)}\right)}
	\end{equation}
	where $d_{fused, x_i}^c$ is the description degree of class $c$ to sample $x_i$, $p_{tar, x_i}^c\left(x_i, \theta_{tar}\right)$ is the prediction probability of sample $x_i$ for class $c$ by the target branch. 
	
	\subsection{The Joint Multi-task Loss}
	
	Combining the above modules, we get a multi-task FER framework, as shown in Figure \ref{fig:model}. By optimizing label distribution generation and facial expression prediction together, the model's performance is advanced. The overall loss is formulated as follows:
	
	\begin{equation}
		L = L_{RR} + \alpha_1 \cdot L_{ce} + \alpha_2 \cdot L_{kld}
	\end{equation}
	with
	\begin{gather}
		\alpha_1 = 
		\begin{cases}
			1 & e \le \beta \\
			\exp{\left(-\left(1-\frac{\beta}{e}\right)^2\right)} & e > \beta \\
		\end{cases} \\
		\alpha_2 = 
		\begin{cases}
			\exp{\left(-\left(1-\frac{e}{\beta}\right)^2\right)} & e \le \beta \\
			1 & e > \beta 
		\end{cases}
	\end{gather}
	
	where $\alpha_1$ and $\alpha_2$ are the weighted ramp functions w.r.t the epoch $e$, and $\beta$ is the epoch threshold. In the first few epochs, we focus on training the auxiliary branch to ensure that it robustly outputs both label distributions and class distributions. In the latter epochs, we shift our focus to training the target branch and avoid overfitting of the auxiliary branch. During the inference phase, the auxiliary branch is removed, and only the target branch is used to predict the expressions of the samples. Therefore, our framework is an end-to-end framework and incurs no additional training cost. 
	
	\section{Experiments}
	\label{sec:exp}
	\subsection{Datasets}
	
	RAF-DB \citep{li2017reliable} contains 29672 real world images obtained from the internet which are splited into single-label dataset and multi-label dataset. We use its single-label subset which has a train set of 12271 images and a test set of 2478 images. The single-label subset is annotated with 7 basic expressions (i.e. neutral, happiness, surprise, sadness, anger, disgust and fear).
	
	AffectNet \citep{mollahosseini2017affectnet} collects over 1 million real word images online, but only 291650 images are labeled with the 8 basic expressions labels and the Arousal/Valence values. Its training set has 287651 images and its test set has 3999 images. We will only use samples that are labeled with 7 basic expressions labels. The samples labeled with contempt are abandoned for fair comparison. 
	
	SFEW \citep{yu2015image} are composed of static frames chosen from the AFEW dataset based on facial point clustering. SFEW has a training set of 958 images, a validation set of 436 images and a test set of 272 images, which are all labeled with the 7 basic expression label. As the label list of the training set are not public, we measure our model performance on its validation dataset by convention.
	
	\subsection{Experimental Details}
	
	\textbf{Training Details}: We train our Ada-DF on CUDA 10.2, PyTorch 1.12.0, torchvision 0.13.0 and Python 3.9 with a Tesla V100 GPU. We use MTCNN to detect faces and locate their landmark points in RAF-DB, AffectNet and SFEW, and then align them via similarity transformation. All images are further resized to 100 $\times$ 100 before training. The backbone network is ResNet18 by default, which is pre-trained on the MS-Celeb-1M dataset for face recognition. RandAugment \citep{cubuk2020randaugment}, random horizontal flipping and random erasing \citep{zhong2020random} are used for data augmentation in an online manner. Specially, SFEW makes additional use of TenCrop strategy for its lack of examples. The batch size for RAF-DB and AffectNet is set as 64 and the batch size for SFEW is set as 16 due to its lack of samples. The learning rate is initialized to 0.001. We use Adam optimizer and ExponentialLR learning rate scheduler with the gamma of 0.9 to decay the learning rate after each epoch. The training process ends at epoch 75. 
	
	\textbf{Evaluation Metric}: The most used metric in FER is total accuracy, so we record our experimental results with the evaluation indicator accuracy $Acc$, which is computed as follows:
	
	\begin{equation}
		Acc = \frac{1}{N} \sum_{1}^{N} I\left(y_i == \hat{y}_i\right)
	\end{equation}
	
	where $I\left(\cdot\right)$ is the indicator function, $y_i$ and $\hat{y}_i$ are the true label and the predicted label of the $i$-th samples $x_i$ respectively.

	\subsection{Parameter Analysis}
	

			\begin{table}
				\centering	
				\caption{Ablation study on hyper-parameters}
				\label{table:ablation}	
				\begin{subtable}{\linewidth}
						\centering
						\setlength{\tabcolsep}{1.0mm}
						\setlength{\abovecaptionskip}{0cm}
						\setlength{\belowcaptionskip}{-0.2cm}
						\caption{Performance of different minimum weights $w_{min}$}
						\begin{tabular}{cccccc}
								\toprule
								$w_{min}$ & 0.0 & 0.1 & 0.2 & 0.3 & 0.4 \\
								\midrule
								RAF-DB & 89.55\% & 89.85\% & \textbf{90.04\%} & 89.80\% & 89.88\% \\
								AffectNet & 65.26\% & 65.13\% & \textbf{65.34\%} & 65.13\% & 65.17\% \\
								SFEW & 59.36\% & 59.95\% & \textbf{60.46\%} & 59.01\% & 58.99\% \\
								\bottomrule
							\end{tabular}
					\end{subtable}
				\begin{subtable}{\linewidth}
						\centering
						\setlength{\tabcolsep}{1.0mm}
						\setlength{\abovecaptionskip}{1.0mm}
						\setlength{\belowcaptionskip}{1.0mm}
						\caption{Performance of different distribution thresholds $t$}
						\begin{tabular}{cccccc}
								\toprule
								$t$ & 0.4 & 0.5 & 0.6 & 0.7 & 0.8 \\
								\midrule
								RAF-DB & 89.71\% & 89.83\% & 89.96\% & \textbf{90.04\%} & 89.80\% \\
								AffectNet & 65.14\% & \textbf{65.34\%} & 65.12\% & 65.07\% & 65.22\% \\
								SFEW & 57.80\% & \textbf{60.46\%} & 41.10\% & 37.25\% & 23.12\% \\
								\bottomrule
							\end{tabular}
					\end{subtable}
				\begin{subtable}{\linewidth}
						\centering
						\setlength{\tabcolsep}{1.0mm}
						\setlength{\abovecaptionskip}{1.0mm}
						\setlength{\belowcaptionskip}{1.0mm}
						\caption{Performance of different epoch thresholds $\beta$}
						\begin{tabular}{cccccc}
								\toprule
								$\beta$ & 1 & 3 & 5 & 7 & 9 \\
								\midrule
								RAF-DB & 89.91\% & \textbf{90.04\%} & 89.83\% & 88.95\% & 89.72\% \\
								AffectNet & 65.16\% & 65.30\% & \textbf{65.34\%} & 65.00\% & 65.12\% \\
								SFEW & 57.85\% & 56.29\% & \textbf{60.46\%} & 59.72\% & 59.34\% \\
								\bottomrule
							\end{tabular}
					\end{subtable}
			\end{table}
	
	In this section, we aim to explore the different contributions of different hyper-parameters using ResNet18 as backbone.
	
	\textbf{Analysis of minimum weight $\bm{w_{min}}$}: The minimum weight $w_{min}$ is used to control the degree of distribution fusion. For samples with low attention weights, their label distributions extracted from the auxiliary branch are not trustworthy, so we choose to normalize the attention weights to between $w_{min}$ and 1 in order to fuse more class distributions in the final fused distributions. As shown in Table \ref{table:ablation}, we find the default setting $w_{min} = 0.2$ achieves the best result. When the minimum weight $w_{min}$ is too high, the final distributions will be more like class distributions of emotions which can not describe those ambiguous samples well. When the minimum weight $w_{min}$ is too low, the fused distributions for samples with low weights will be close to label distributions of samples, which contains more biases and could deteriorate the model's performance. 
	

	\textbf{Analysis of distribution threshold $\bm{t}$}: In case that the auxiliary branch can not produce robust label distributions of samples in the first few epochs and fail to obtain the desired class distributions of samples, we set the threshold $t$ to form the threshold distributions when both label and class distributions are not well prepared for training the target branch. The results are shown in Table \ref{table:ablation}. For the dataset RAF-DB, the best accuracy is achieved at $t = 0.7$, while AffectNet and SFEW get the highest accuracy at $t = 0.5$. For relatively certain and professionally annotated datasets like RAF-DB, the corresponding description degree of each emotion is high because less ambiguous samples lead to less uncertainty in the final distributions, thus a higher distribution threshold $t$ can train the auxiliary branch better. But for datasets containing more uncertain and ambiguous samples such as AffectNet and SFEW, the samples in datasets tend to contain more emotions, making description degrees of other categories higher and the description degree of corresponding emotion lower, so threshold distributions with a lower distribution threshold $t$ can fit the ground-truth distributions better.

	
	\textbf{Analysis of epoch threshold $\bm{\beta}$}: The parameter $\beta$ decides how fast the auxiliary and target branches are trained. As shown in Table \ref{table:ablation}, the best results are achieved when $\beta$ is 3, 5 and 5 respectively. It is intuitive that for datasets with more uncertain samples, we would better train the auxiliary branch with more effort to extract more trustworthy label distributions in the first few epochs, which provide more robust supervision for the target branch in the following epochs.
	
	\subsection{Distribution Analysis}

	\begin{table}
		\setlength{\tabcolsep}{0.8mm}
		\setlength{\abovecaptionskip}{0cm}
		\setlength{\belowcaptionskip}{-0.2cm}
		\centering
		\caption{Ablation study on different distributions}
		\label{table:ablation1}
		\begin{tabular}{ccccccc}
			\toprule
			$\mathbf{y}$ & $\mathbf{d}_{class}$ & $\mathbf{d}_{label}$ & $\mathbf{d}_{fused}$ & RAF-DB & AffectNet & SFEW \\
			\midrule
			\checkmark &  &  &  & 88.71 & 64.58 & 53.90 \\
			\checkmark & \checkmark &  &  & 89.28 & 64.63 & 56.05 \\
			\checkmark &  & \checkmark &  & 89.50 & 64.95 & 60.23 \\
			\checkmark & \checkmark & \checkmark & \checkmark & \textbf{90.04} & \textbf{65.34} & \textbf{60.46} \\
			\bottomrule
		\end{tabular}
	\end{table}
	
	To separately evaluate the roles of different modules and distributions, we conduct extensive ablation studies on RAF-DB, AffectNet, and SFEW using ResNet18 as backbone. The results are shown in Table \ref{table:ablation1}.
	
	\textbf{Label Distribution Extraction}: The label distributions of samples $\mathbf{d}_{label}$ are directly extracted from the auxiliary branch, which facilitate the training of the target branch on three datasets. Compared to the baseline (single-label learning on ResNet18), the accuracy is advanced by 0.79\%, 0.37\%, and 6.33\%. It indicated that the information implied by label distributions effectively avoids the overfitting and enables the model to better learn ambiguous samples. 
	
	\textbf{Class Distribution Mining}: The class distributions of emotions $\mathbf{d}_{class}$ are obtained by simply averaging the label distributions of corresponding emotions respectively. By replacing $\mathbf{d}_{label}$ with $\mathbf{d}_{class}$, the accuracy of Ada-DF becomes worse, but still superior to the baseline. Assuming the biases in the label distributions of samples follow the normal distribution, the class distribution mining not only eliminates those biases, but also the diversity of samples, thus making it hard to avoid intra-class variations. 
	
	\textbf{Adaptive Distribution Fusion}: To utilize both label distributions of samples $\mathbf{d}_{label}$ and class distributions of emotions $\mathbf{d}_{class}$, we finally adaptively fuse two distributions bases on attention weights to obtain more precise distributions. We can find that the final accuracies significantly exceed the baseline by 1.33\%, 0.76\% and 6.56\% respectively, which reflects that the fused distributions better fit with samples' actual distributions than either label or class distributions. 
	
			\begin{table}
		\centering
		\begin{threeparttable}
			\setlength{\tabcolsep}{2.4mm}
			\setlength{\abovecaptionskip}{0cm}
			\setlength{\belowcaptionskip}{-0.2cm}
			\centering
			\caption{Comparison with the state of the art}
			\label{table:compare}
			\begin{tabular}{cccc}
				\toprule
				Method & RAF-DB & AffectNet & SFEW \\ 
				\midrule
				LDL-ALSG \citep{chen2020label} \tnote{$\ast$} & 85.53 & / & 56.5 \\
				SCN \citep{wang2020suppressing} \tnote{$\ast$} & 87.03 & / & / \\
				DDL \citep{ruan2020deep} & 87.71 & / & 59.86 \\
				OADN \citep{ding2020occlusion} & 89.83 & 64.06 & / \\
				KTN \citep{li2021adaptively} & 88.07 & 63.97 & / \\
				DMUE \citep{she2021dive} \tnote{$\ast$} & 89.42  & / & 58.34 \\
				PT \citep{jiang2021boosting} & 89.57 & / & / \\
				LAENet-SA \citep{wang2021light} \tnote{$\ast$} & / & 64.09 & / \\
				PAT \citep{cai2022probabilistic} & 88.43 & / & 57.57 \\
				IPD-FER \citep{jiang2022disentangling} & 88.89 & 62.23 & 58.43 \\
				EAC \citep{zhang2022learn} & 90.35 & 65.32 & / \\
				ReSup \citep{zhang2023resup} & 89.70 & 65.46 & / \\
				ContraWarping \citep{xue2023unsupervised} & 89.18 & 64.94 & /\\
				\textbf{Ours (ResNet18)} & 90.04 & 65.34 & \textbf{60.46} \\ 
				\textbf{Ours (ResNet50)} & \textbf{90.35} & \textbf{65.68} & 20.32 \\ 
				\bottomrule
			\end{tabular}
			\begin{tablenotes}
				\footnotesize
				\item [$\ast$] These methods have results on AffectNet of 8 labels, which we do not compare with for our model is trained on AffectNet of 7 labels. 
			\end{tablenotes}
		\end{threeparttable}
	\end{table}

	\subsection{Performance Comparison}
	
	We compare our Ada-DF with existing state-of-the-art methods to measure the model performance. As shown in Table \ref{table:compare}, we outperform the previous works and achieve the best results on all datasets. Our method is evaluated on ResNet18 and ResNet50 respectively.

	Most FER methods of recent years are single-label learning methods. Since single labels lack rich sentiment information, single-label enhances expression feature extraction by applying attention mechanism and excluding expression-independent distractors such as occlusion, imbalanced data, sample identity, etc.
	
	Our method is a label distribution learning method, which can provide rich supervision for training the model than single-label learning methods, thus achieving superior performance comparing to single-label learning methods. Nevertheless, few FER methods utilize label distribution learning as most FER datasets only provide single labels. We select two typical LDL methods for comparison and discussion. DMUE generates the label distributions of samples by constructing multiple auxiliary branches which corresponds to different classes of expressions respectively, excluding interference from samples of different classes. However, the multi-branch framework of DMUE consumes much time and space, and cannot omit mislabeled samples in the dataset. LDL-ALSG improves the performance by minimizing the distance between the logical label and the label distribution of a sample and the distance between the label distribution of a sample and the label distributions of its neighbors in the label space of action unit recognition and facial landmark detection. However, LDL-ALSG requires to generate action units and facial landmarks of samples before training, which make it not an end-to-end method. In contrast, our method only constructs one auxiliary branch and obtains accurate distributions of samples via the adaptive distribution fusion module without introducing extra information, showing superior performance over other LDL methods.
	
	Specially, Ada-DF achieves better results on RAF-DB and AffectNet using ResNet50 compared with ResNet18, but the accuracy of SFEW drops a lot as for a relatively bigger model ResNet50 SFEW contains too few samples to generate enough robust distributions in the first few epochs, thus deteriorating the overall performance, so it is better to train the dataset containing much fewer samples with a smaller model.

	\subsection{Evaluation on Synthetic Ambiguity}
	We evaluate our proposed Ada-DF under different levels of label noises (10\%, 20\%, 30\%) on RAF-DB to demonstrate its effectiveness using ResNet18 and ResNet50 as the backbone respectively. The label noises are generated following the previous works \citep{she2021dive, wang2020suppressing}.
	
	The results, as shown in Table \ref{table:noise}, demonstrate the superior performance of our Ada-DF compared to other state-of-the-art noisy label learning methods. For instance, Ada-DF outperforms EAC \citep{zhang2022learn} by 0.93\%, 1.21\% and 1.69\% under 10\%, 20\% and 30\% label noise respectively, which shows the effectiveness of Ada-DF becomes more obvious when the label noise increases. It is worth noting that Ada-DF and EAC achieves the same accuracy on RAF-DB using ResNet50, but for synthetic noise our method could get the similar results just using ResNet18 with fewer parameters, showing the robustness and superiority of Ada-DF against the ambiguity.  

	For single-label learning methods, SCN tries to solve the ambiguity problem by modifying the original labels, which may introduce the risk of changing the correct labels to incorrect labels, whereas Ada-DF automatically tells apart clear samples and noisy samples via the attention module and extracts robust label distributions to avoid the defects of label noises.
		\begin{table}
		\setlength{\tabcolsep}{3.9mm}
		\setlength{\abovecaptionskip}{0cm}
		\setlength{\belowcaptionskip}{-0.2cm}
		\centering
		\caption{Evaluation of Ada-DF on noisy RAF-DB dataset}
		\label{table:noise}
		\begin{tabular}{cccc}
			\toprule
			Methods & Backbone & Noise & RAF-DB\\
			\midrule
			SCN \citep{wang2020suppressing} & ResNet18 & 10 & 82.18\\
			DMUE \citep{she2021dive} & ResNet18 & 10 & 83.19\\
			RUL \citep{zhang2021relative} & ResNet18 & 10 & 86.22\\
			EAC \citep{zhang2022learn} & ResNet50 & 10 & 88.02\\
			\textbf{Ours} & ResNet18 & 10 & 87.81\\
			\textbf{Ours} & ResNet50 & 10 & \textbf{88.95}\\
			\midrule
			SCN \citep{wang2020suppressing} & ResNet18 & 20 & 80.10\\
			DMUE \citep{she2021dive} & ResNet18 & 20 & 81.02\\
			RUL \citep{zhang2021relative} & ResNet18 & 20 & 84.34\\
			EAC \citep{zhang2022learn} & ResNet50 & 20 & 86.05\\
			\textbf{Ours} & ResNet18 & 20 & 86.67\\
			\textbf{Ours} & ResNet50 & 20 & \textbf{87.26}\\
			\midrule
			SCN \citep{wang2020suppressing} & ResNet18 & 30 & 77.46\\
			DMUE \citep{she2021dive} & ResNet18 & 30 & 79.41\\
			RUL \citep{zhang2021relative} & ResNet18 & 30 & 82.06\\
			EAC \citep{zhang2022learn} & ResNet50 & 30 & 84.42\\
			\textbf{Ours} & ResNet18 & 30 & 84.38\\
			\textbf{Ours} & ResNet50 & 30 & \textbf{86.11}\\
			\bottomrule
		\end{tabular}
	\end{table}
	For label distribution learning methods, DMUE introduces a multi-branch framework to extract the label distributions, which can hardly omit the mislabeled samples during training. Our Ada-DF solves it via adaptive distribution fusion to obtain more robust and accurate distributions.

	\subsection{Visualization Analysis}
	
	\subsubsection{Visualization of Features}
	
		\begin{figure*}[t!]
		\centering
		\subfloat[\label{fig:a}]{\includegraphics[width=.5\textwidth]{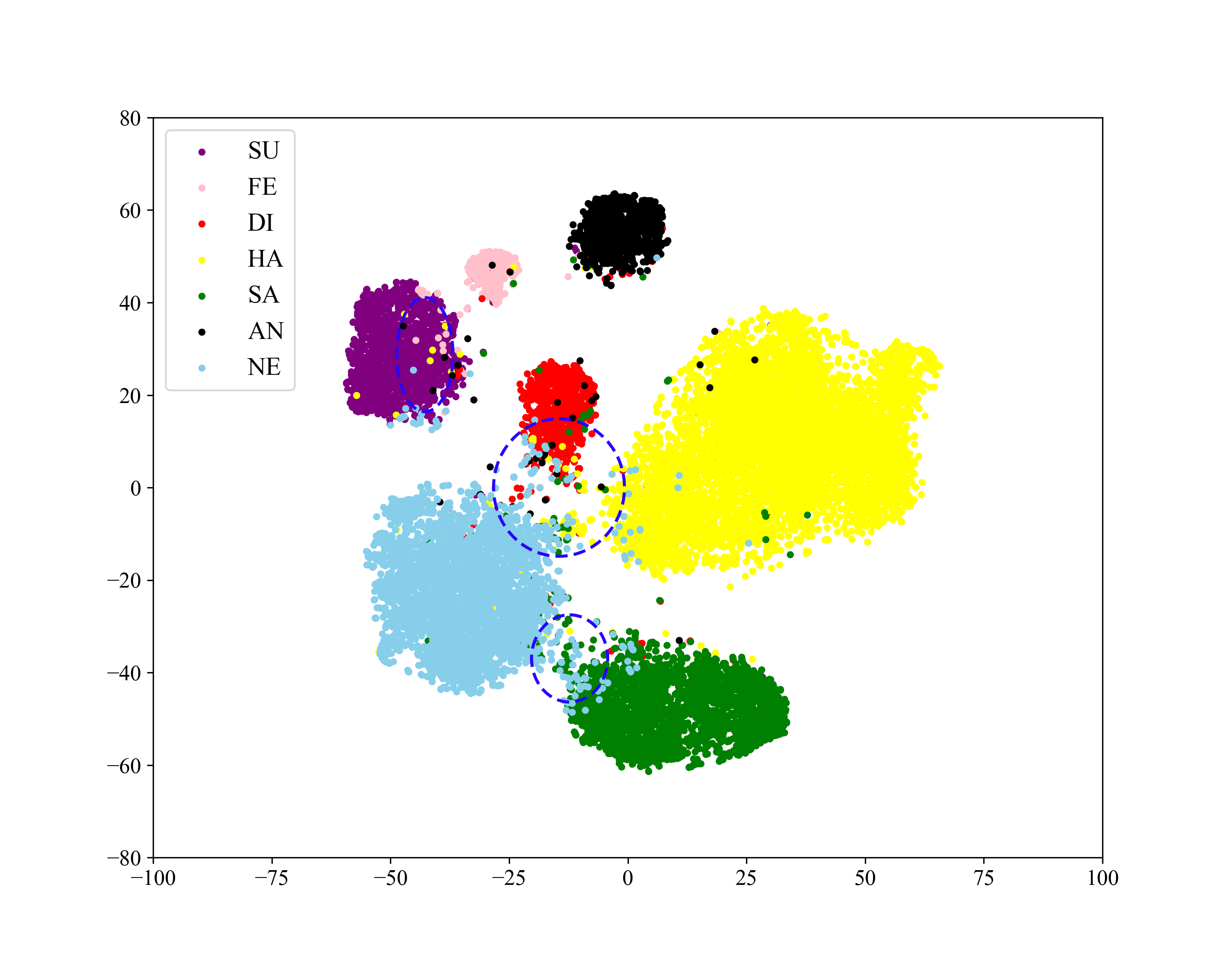}}
		\subfloat[\label{fig:b}]{\includegraphics[width=.5\textwidth]{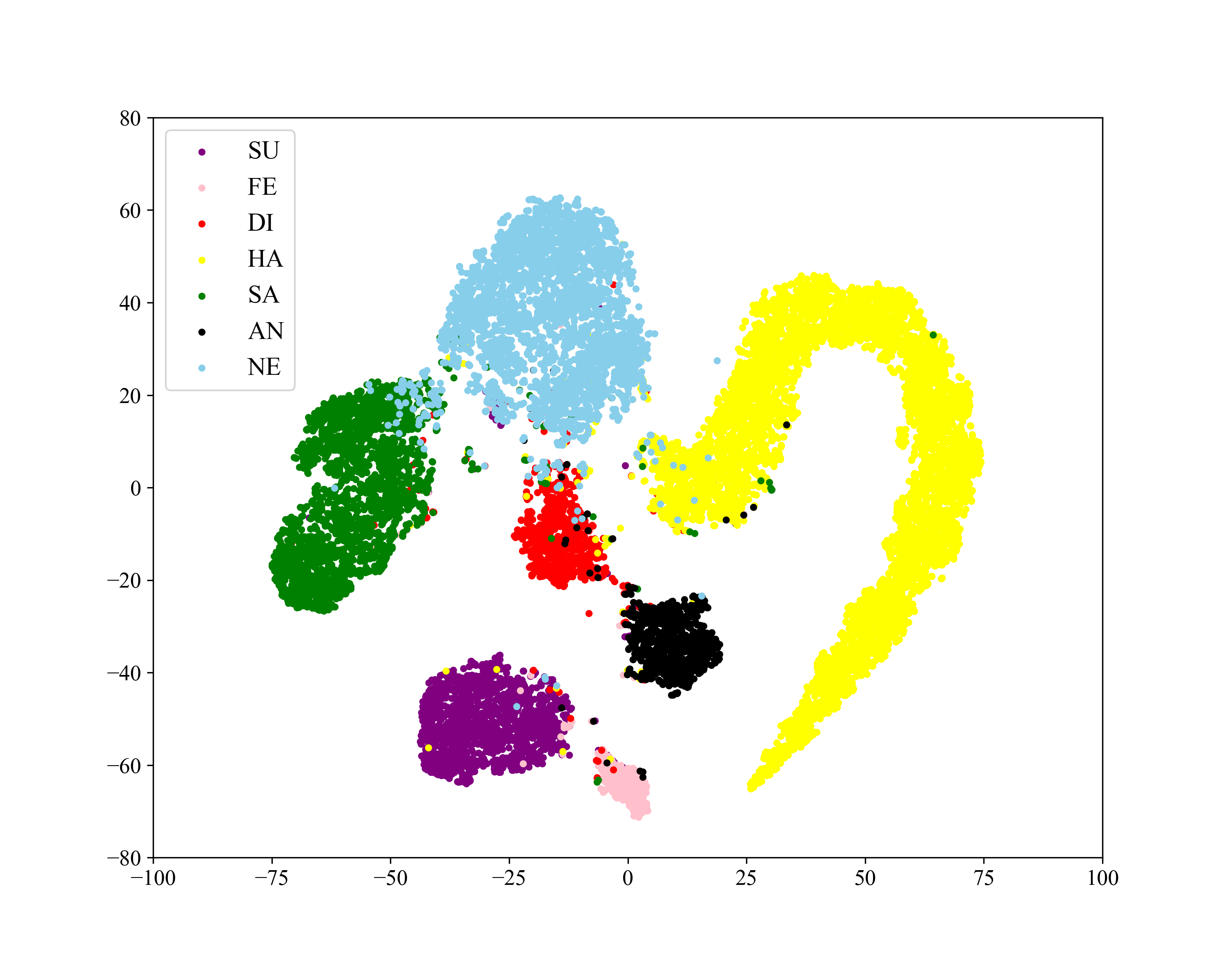}}
		\caption{T-SNE visualization of the outputs in the final hidden layer of the baseline and our approach}
		\label{fig:tsne}
	\end{figure*}

	To better understand our Ada-DF intuitively, we employ t-SNE \citep{van2008visualizing}, a popular dimensionality reduction technique, to plot the learned features on RAF-DB. As shown in Figure \ref{fig:tsne}, the learned features are represented as clusters corresponding to 7 basic expressions. Comparing to baseline, we observe that our method achieves more compact and well-separated clusters. Furthermore, our Ada-DF can achieve a clear boundary between sadness and other categories, while the learned features of baseline are not discriminative enough for some categories. This outcome indicates that our method effectively learns more meaningful features, reducing the intra-class feature differences and increasing the inter-class feature variations. 
	
	\subsubsection{Visualization of Class Distributions}

		\begin{figure*}
		\centering
		\includegraphics[width=\textwidth]{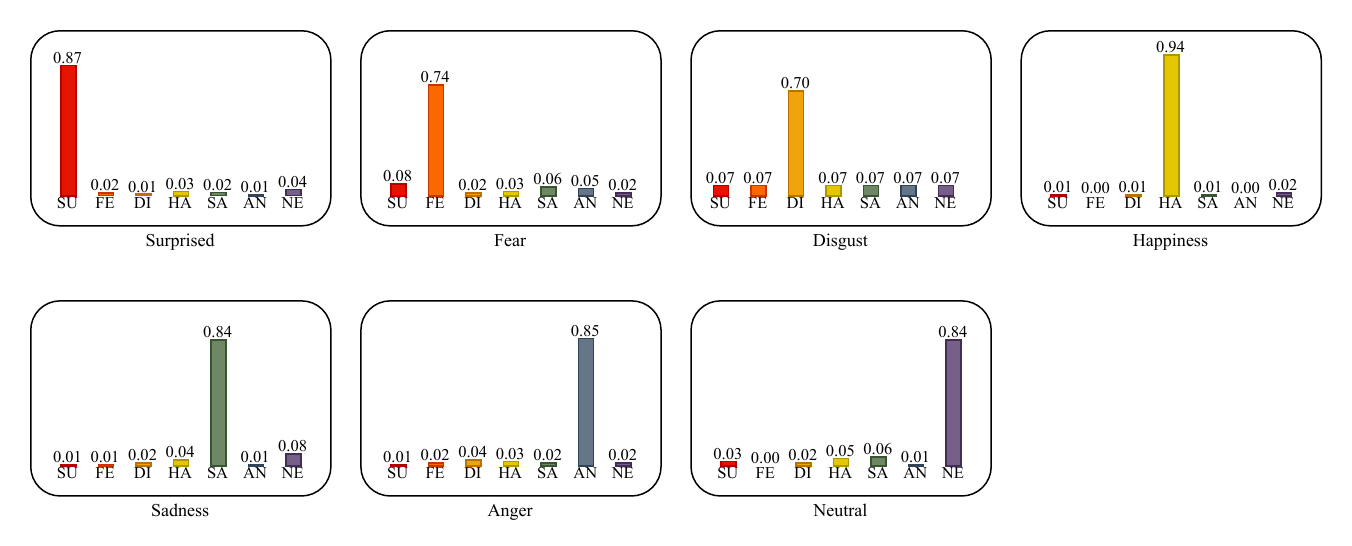}
		\caption{Class distributions of 7 basic emotions mined in RAF-DB. The sum of each distribution may not be equal to 1 due to rounding.}
		\label{fig:clsdis}
	\end{figure*}

	As shown in Figure \ref{fig:clsdis}, we visualize the class distributions of 7 basic emotions mined in RAF-DB to demonstrate how the rich sentiment information and the relations between emotions are found. It's obvious that the description degree of the corresponding emotion for each distribution is the highest, but their values vary. For expressions with large facial muscle movements, such as surprised, happiness and anger, the description degree of the corresponding emotion in the class distribution is higher than that of other emotions. However, the description degree of the corresponding emotion for other emotions are lower because these expressions are more similar and harder to classify.

	\subsubsection{Visualization of Adaptive Distribution Fusion}
	
\begin{figure}
	\centering
	\includegraphics[width=0.8\linewidth]{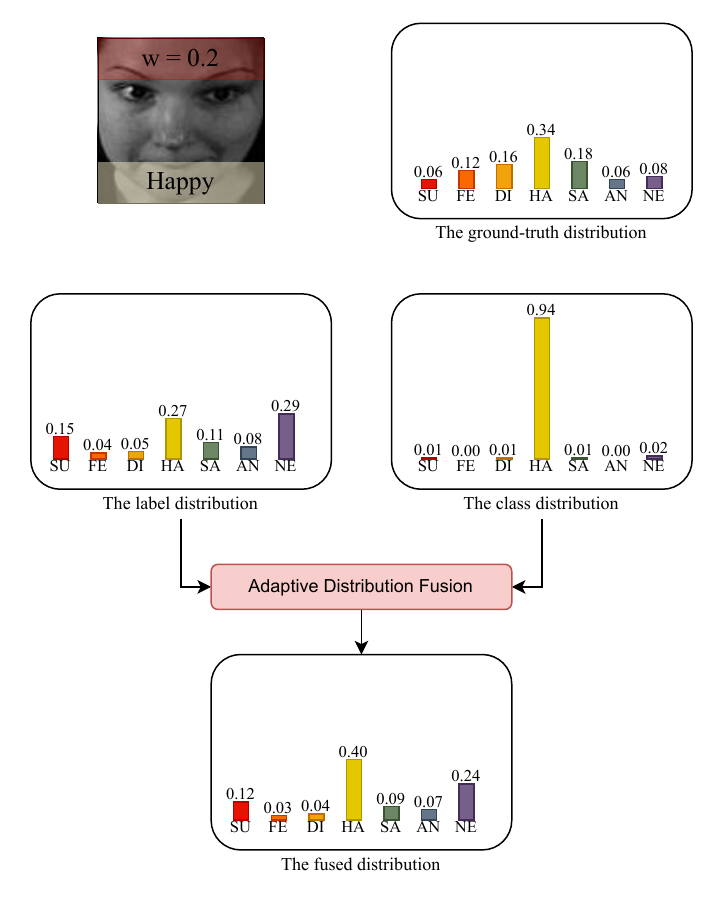}
	\caption{An example of adaptive distribution fusion. The sum of each distribution may not be equal to 1 due to rounding.}
	\label{fig:fusion}
\end{figure}

	We choose the image shown in Figure \ref{fig:example} to show the effectiveness of adaptive distribution fusion, and the distributions are shown in Figure \ref{fig:fusion}. The attention weight of this sample is $0.2$. We consider the probability distribution of annotations by 50 volunteers as the ground-truth distribution for this sample. When comparing the ground-truth distribution with the label distribution of this sample, we observe that the description degree of happy is higher in the ground-truth, while the description degree of neutral is lower. Directly training the model with this label distribution will degrade the model's predictive performance. By utilizing adaptive distribution fusion, we can effectively supplement the description degree of happy and suppress the description degree of neutral in the fused distribution which is more closer to the ground-truth distribution, demonstrating the effectiveness of this step.

	\subsubsection{Visualization of Multi-epoch Distributions}
\begin{figure*}
		\centering
		\includegraphics[width=\textwidth]{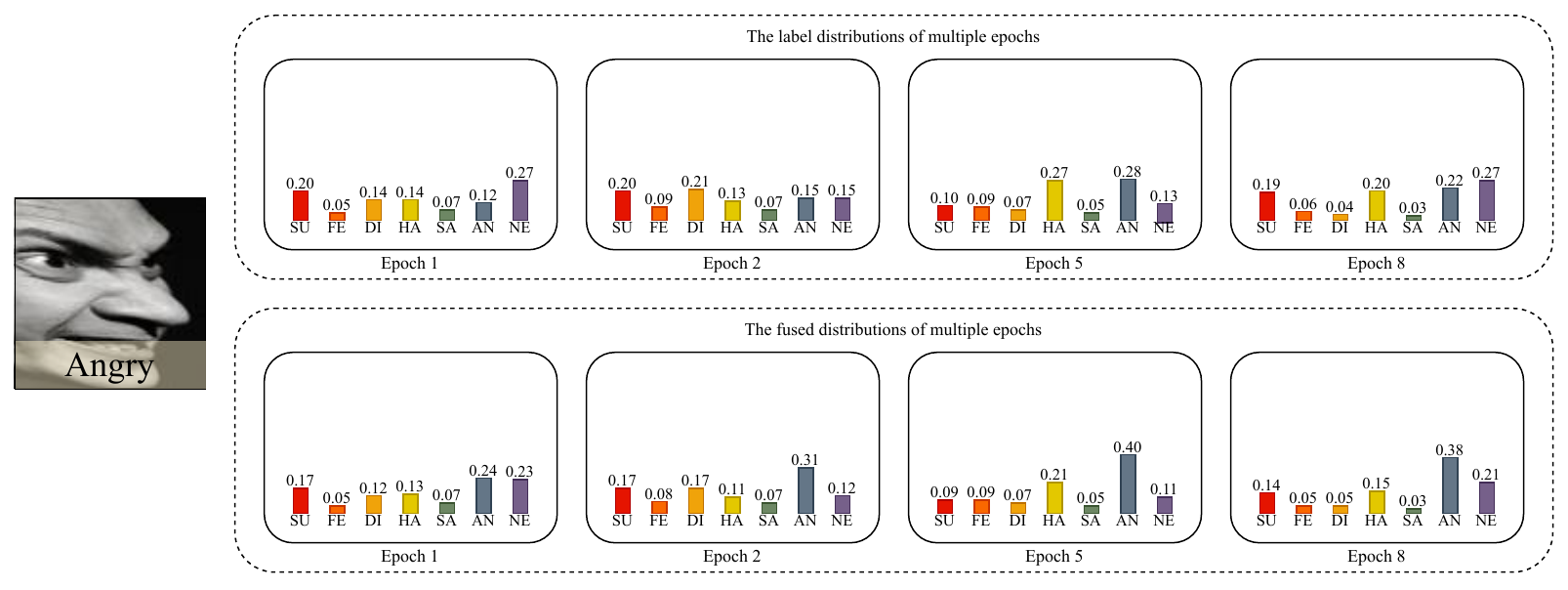}
		\caption{The distributions before and after adaptive distribution fusion of muiltiple epochs.}
		\label{fig:multi}
	\end{figure*}

	To further explain the efficiency of our Ada-DF, we extract the label distributions and the fused distributions of multiple epochs for the sample shown in Figure \ref{fig:multi}. Consistent to our assumption, the label distributions extracted from the auxiliary branch vary in the preliminary epochs. The description degrees of surprise, disgust, happy, fear are tend to be equal, which make this label distribution more like a random output and thus deteriorate the final performance. After adaptive distribution fusion, the description of angry are boosted, which partly eliminate the ambiguity and keep the variety at the same time.

	\section{Conclusion}
	\label{sec:conc}
	
	In this paper, we propose a novel multi-task framework that integrates label distribution generation as an auxiliary task for FER. Our framework comprises an auxiliary branch responsible for label distribution extraction and a target branch for facial expression classification. By extracting label distributions, we are able to mine the class distributions of emotions. These distributions are then adaptively fused, leveraging the strengths of both distributions and obtaining precise fused distributions that provide more accurate supervision for model training. Through extensive experiments on RAF-DB, AffectNet, and SFEW, we demonstrate the effectiveness and robustness of our method. Detailed analysis of different modules and distributions reveals the significant contribution of Label Distribution Extraction. Additionally, the inclusion of Class Distribution Mining and Adaptive Distribution Fusion further improves the accuracy of our method. Our approach successfully applies label distribution generation in FER and has the potential for broader applicability in other deep-learning based tasks.
	
	For future work, we intend to explore the generation of more robust distributions by incorporating more additional FER-related tasks such as facial landmarks detection and facial action units detection. Additionally, we aim to focus on extracting more representative features using multiple modalities, including 3D face images, audio, and other relevant sources of information. These efforts will contribute to advancing the field of FER and expanding the capabilities of our proposed method.
	
	
	
	\vfill\pagebreak
	
	\bibliographystyle{elsarticle-num}
	\bibliography{refs}
	
	
	%
	%
	%
\end{document}